\documentclass[12pt,journal,onecolumn]{IEEEtran}
\usepackage{amsmath,graphicx}
\usepackage[linesnumbered,ruled]{algorithm2e}
\usepackage{xcolor,colortbl}
\usepackage{setspace}
\usepackage[utf8]{inputenc}

\title{Learning First-to-Spike Policies for Neuromorphic Control Using Policy Gradients}
\author{
    \IEEEauthorblockN{Bleema~Rosenfeld\IEEEauthorrefmark{1}, Osvaldo~Simeone\IEEEauthorrefmark{2}, and~Bipin~Rajendran\IEEEauthorrefmark{1}}%
    \\~\\\IEEEauthorblockA{\IEEEauthorrefmark{1}Department of Electrical and Computer Engineering
    \\New Jersey Institute of Technology, Newark, NJ 07102, USA}%
    \\\IEEEauthorblockA{\IEEEauthorrefmark{2}Centre for Telecommunications Research, Department of Informatics,
    \\King’s College London, London, WC2R 2LS, UK.}%
\thanks{O. Simeone is on leave from NJIT}%
\thanks{This research was supported in part by the European Research Council (ERC) under the European
Union’s Horizon 2020 Research and Innovation Program (Grant Agreement No. 725731) and the U.S. National Science Foundation through grants 1525629 and 1710009.}}%

\begin{document}

\maketitle

\begin{abstract}
Artificial Neural Networks (ANNs) are currently being used as function approximators in many state-of-the-art Reinforcement Learning (RL) algorithms. Spiking Neural Networks (SNNs) have been shown to drastically reduce the energy consumption of ANNs by encoding information in sparse temporal binary spike streams, hence emulating the communication mechanism of biological neurons. Due to their low energy consumption, SNNs are considered to be important candidates as co-processors to be implemented in mobile devices. In this work, the use of SNNs as stochastic policies is explored under an energy-efficient first-to-spike action rule, whereby the action taken by the RL agent is determined by the occurrence of the first spike among the output neurons. A policy gradient-based algorithm is derived considering a Generalized Linear Model (GLM) for spiking neurons. Experimental results demonstrate the capability of online trained SNNs as stochastic policies to gracefully trade energy consumption, as measured by the number of spikes, and control performance. Significant gains are shown as compared to the standard approach of converting an offline trained ANN into an SNN. 
\end{abstract}
\begin{IEEEkeywords}
Spiking Neural Network, Reinforcement Learning, Policy Gradient, Neuromorphic Computing.
\end{IEEEkeywords}

\section{Introduction}
\label{sec:intro}
Artificial neural networks (ANNs) are used as parameterized non-linear models that serve as inductive bias for a large number of machine learning tasks, including notable applications of Reinforcement Learning (RL) to control problems \cite{jaderberg2018human}. While ANNs rely on clocked floating- or fixed-point operations on real numbers, Spiking Neural Networks (SNNs) operate in an event-driven fashion on spiking synaptic signals (see Fig. 1). Due to their lower energy consumption when implemented on specialized hardware, SNNs are emerging as an important alternative to ANNs that is backed by major technology companies, including IBM and Intel  \cite{truenorth,davies2018loihi}. Specifically, SNNs are considered to be important candidates as co-processors to be implemented in battery-limited mobile devices (see, e.g., \cite{chen2017machine}). Applications of SNNs, and of associated neuromorphic hardware, to supervised, unsupervised, and RL problems have been reported in a number of works, first in the computational neuroscience literature and more recently in the context of machine learning \cite{rezende2011variational,kappel2018dynamic,jin2018hybrid}. 

\begin{figure}
\centering
\resizebox{0.75\linewidth}{!}{
\begin{minipage}[b]{0.4\linewidth}
  \centering
  \centerline{\includegraphics[width=\linewidth,trim={2.35in 6in 3in 2in},clip]{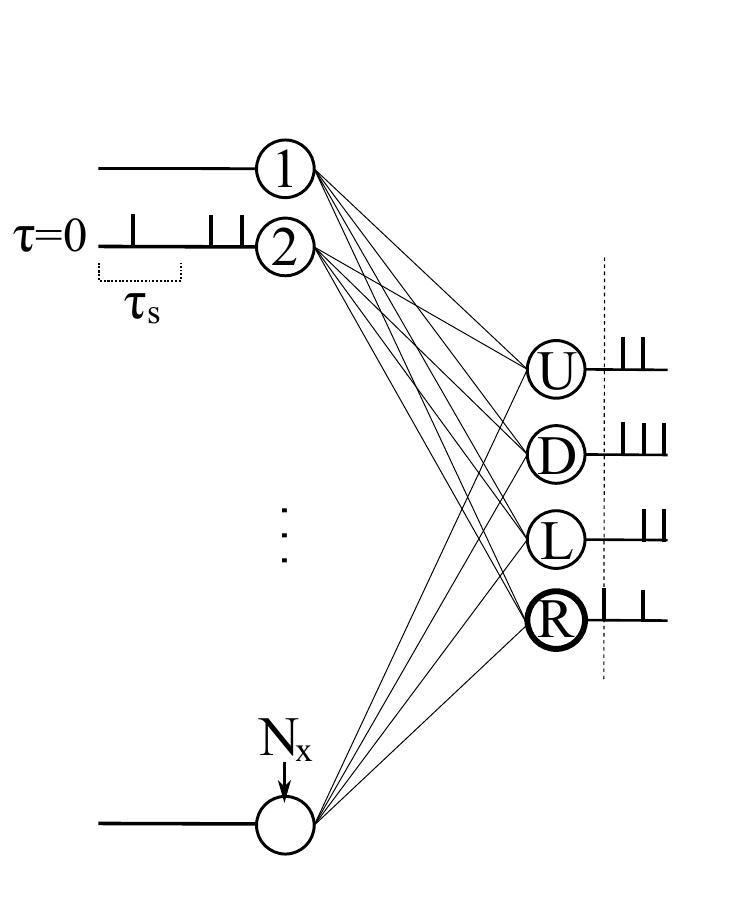}}
  \centerline{(a)}\medskip
\end{minipage}
\hfill   
\begin{minipage}[b] {0.58\linewidth}
    \centering
    \centerline{\includegraphics[width=\linewidth,trim={1in 6.2in 1in 0in},clip]{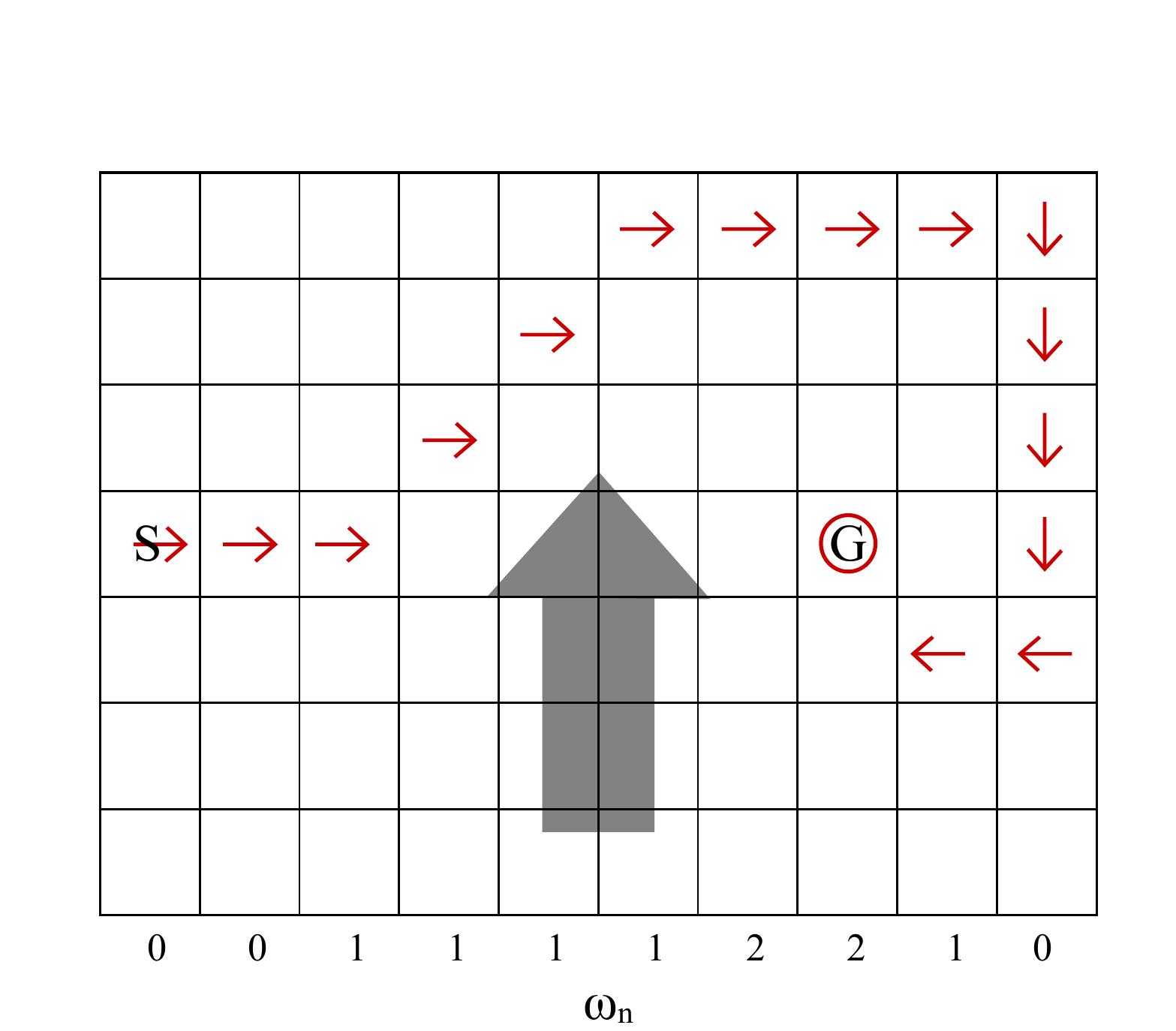}}
    \centerline{(b)}\medskip
\end{minipage}}
\caption{(a) SNN first-to-spike policy with action selected (R in the illustration) among Up, Down, Left, and Right marked with a bold line and decision time marked with a dashed vertical line; (b) An example of a realization of an action sequence in a windy grid-world problem.} 
\label{fig:optimal policy}
\end{figure}

SNN models can be broadly classified as \emph{deterministic}, such as the leaky integrate-and-fire (LIF) model \cite{gerstner2002spiking} or \emph{probabilistic}, such as the generalized linear model (GLM) \cite{pillow2005prediction}. Prior work on RL using SNNs has by and large adopted deterministic SNN models to define action-value function approximators. This is typically done by leveraging \emph{rate decoding} and either \emph{rate encoding} \cite{zheng2017hardware,nakano2015spiking}, or time encoding \cite{bing2018end}. Under rate encoding and decoding, the spiking rates of input and output neurons represent the information being processed and produced, respectively, by the SNN. A standard approach, to be considered here as baseline, is to train an ANN offline and to then convert the resulting policy into a deterministic SNN with the aim of ensuring that the output spiking rates are close to the numerical output values of the trained ANN \cite{diehl2015fast,nakano2015spiking}. There is also significant work in the theoretical neuroscience literature concerning the definition of biologically plausible online learning rules \cite{florian2007reinforcement, shim2017biologically,vasilaki2009spike}. 

In all of the reviewed studies, exploration is made possible by a range of mechanisms such as $\epsilon$-greedy in \cite{shim2017biologically} and stochasticity introduced at the synaptic level \cite{nakano2015spiking,vasilaki2009spike}, requiring the addition of some external source of randomness. As a related work, reference \cite{zheng2017hardware} discusses the addition of noise to a deterministic SNN model to induce exploration of the state space from a hardware perspective. In contrast, in this paper, we investigate the use of probabilistic SNN policies that naturally enable exploration thanks to the inherent randomness of their decisions, hence making it possible to learn while acting in an on-policy fashion.

Due to an SNN's event-driven activity, its energy consumption depends mostly on the number of spikes that are output by its neurons. This is because the idle energy consumption of neuromorphic chips is generally extremely low (see, e.g., \cite{truenorth,davies2018loihi}). With this observation in mind, this work proposes the use of a probabilistic SNN, based on GLM spiking neurons, as a stochastic policy that operates according to a first-to-spike decoding rule. The rule outputs a decision as soon as one of the output neurons generates a spike, as illustrated in Fig. 1(a), hence potentially reducing energy consumption. A gradient-based updating rule is derived that leverages the analytical tractability of the first-to-spike decision criterion under the GLM model. We refer to  \cite{bagheri2017training} for an application of the first-to-spike rule to a supervised learning classification algorithm. 

The rest of the paper is organized as follows. Sec. \ref{sec:prob def} describes the problem formulation and the GLM-based SNN model. Sec. \ref{sec:polgrad ftssnn} introduces the proposed policy gradient on-policy learning rule. Sec. \ref{sec:baselineSNN} reviews the baseline approach of converting an offline trained ANN into an SNN. Experiments and discussions are reported in Sec. \ref{sec:results}.  

\section{Problem Definition and Models}
\label{sec:prob def}
\textbf{Problem definition.} We consider a standard RL single-agent problem formulated on the basis of discrete-time Markov Decision Processes (MDPs). Accordingly, at every time-step $t=1,2,...$, the RL agent takes an action $A_t$ from a finite set of options based on its knowledge of the current environment state $S_t$ with the aim of maximizing a long-term performance criterion. The agent's policy $\pi(A|S,\theta)$ is a parameterized probabilistic mapping from the state space to the action space, where $\theta$ is the vector of trainable parameters. After the agent takes an action $A_t$, the environment transitions to a new state $S_{t+1}$ which is observed by the agent along with an associated numeric reward signal $R_{t=1}$, where both $S_{t+1}$ and $R_{t+1}$ are generally random functions of the state $S_t$ and action $A_t$ with unknown conditional probability distributions. 

An episode, starting at time $t=0$ in some state $S_0$, ends at time $t^{\text{G}}$, when the agent reaches a goal state $S^{\text{G}}$. The performance of the agent's policy $\pi$ is measured by the long-term discounted average reward
\begin{align}\label{eq:value}
    V_{\pi}(S_0)=\sum_{t=0}^{\infty}\gamma^{t}\text{E}_\pi [R_{t}], 
\end{align}where $0<\gamma<1$ is a discount factor. The reward $R_t$ is assumed to be zero for all times $t>t^{\text{G}}$. With a proper definition of the reward signal $R_t$, this formulation ensures that the agent is incentivized to reach the terminal state in as few time-steps as possible. 

While the approach proposed in this work can apply to arbitrary RL problems with discrete finite action space, we will focus on a standard windy grid-world environment \cite{sutton1998reinforcement}. Accordingly, as seen in Fig. 1(b), the state space is an $M \times N$ grid of positions and the action space is the set of allowed horizontal and vertical single-position moves, i.e., Up, Down, Left, or Right. The start state $S_0$ and the terminal state $S^{\text{G}}$ are fixed but unknown to the agent. Each column $n=1,..,N$ in the grid is subject to some unknown degree of `wind', which pushes the agent upward by $\omega_n$ spaces when it moves from a location in that column. The reward signal is defined as $R_{t+1}>0\text{ if }S_{t+1}=S^{\text{G}}$ and, otherwise, we have $R_{t+1}=0$. 

\textbf{Probabilistic SNN model.} In order to model the policy $\pi(A|S,\theta)$, as we will detail in the next section, we adopt a probabilistic SNN model. Here we briefly review the operation of GLM spiking neurons \cite{pillow2005prediction}. Spiking neurons operate over discrete time $\tau=1,...,T$ and output either a ``0'' or a ``1'' value at each time, where the latter represents a spike. We emphasize that, as it will be further discussed in the next section, the notion of time $\tau$ for a spiking neuron is distinct from the time axis $t$ over which the agent operates. Consider a GLM neuron $j$ connected to $N_s$ pre-synaptic (input) neurons. At each time instant $\tau=1,...,T$ of the neuron's operation, the probability of an output spike at neuron $j$ is given as $\sigma(u_{j,\tau})$, where $\sigma(x)=1/(1+\text{exp}(-x))$ is the sigmoid function and $u_{j,\tau}$ is the neuron's membrane potential\begin{align}\label{eq:pot}
    u_{j,\tau} = \sum_{i=1}^{N_s}\alpha_{i,j}^{\dagger} x_{i,\tau-\tau_s:\tau-1}+b_j.
\end{align}In (\ref{eq:pot}), the $\tau_s \times 1$ vector $\alpha_{i,j}$ is the so called \textit{synaptic kernel} which describes the operation of the synapse from neuron $i$ to neuron $j$ with $\dagger$ defined as the transpose operator here; $b_j$ is a bias parameter; and $x_{i,\tau-\tau_s:\tau-1}$ collects the past $\tau_s$ samples of the $i$th input neuron. As in \cite{pillow2005prediction}, we model the synaptic kernel as a linear combination $\alpha_{i,j}=Bw_{i,j}$ of $K_s$ basis functions, described by the columns of $\tau_s\times K_s$ matrix $B$, with the $K_s\times 1$ weight vector $w_{i,j}$. We specifically adopt the raised cosine basis functions in \cite{pillow2005prediction}.
\begin{algorithm}[!t]
\DontPrintSemicolon
\label{alg:ftsreinforce}
    \SetKwInOut{Input}{Input}
    \Input{randomly initialized parameter $\theta$, learning rate $\eta_i,\ i=1,2,...$}
    $i=1$\;
    \Repeat{convergence}{
    \While{$S_t \neq S^\text{G}$}{
      encode $S_t$ in spike domain\;
      run SNN and set $A_t\leftarrow \text{index of FtS neuron}$\;
      \text{observe next state and reward} $S_{t+1},R_{t+1}$\;
    }
    $V_{t^{\text{G}}+1}=0$\;
    \For{t=$t^{\text{G}}:-1:1$}{
    $V_t = R_{t+1}+\gamma V_{t+1}$\;
    $\theta \leftarrow \theta + \eta_i \nabla_{\theta}\log\pi(A_t|S_t,\theta)V_t$\;
    }
    $i\leftarrow i+1$\;
    }
    \caption{Policy Gradient Rule for First-to-Spike (FtS) SNNs}
\end{algorithm}

\section{Policy-Gradient Learning Using First-to-Spike SNN Rule}
\label{sec:polgrad ftssnn}

In this section, we propose an on-policy learning algorithm for RL that uses a first-to-spike SNN as a stochastic random policy. Although the approach can be generalized, we focus here on the fully connected two-layer SNN shown in Fig. \ref{fig:optimal policy}(a). In the SNN, the first layer of $N_x$ neurons encodes the current state of the agent $S_t$, as detailed below, while there is one output GLM neuron for every possible action $A_t$ of the agent, with $N_s=N_x$ inputs each. For example, in the grid world of Fig. \ref{fig:optimal policy}(b), there are four output neurons. The policy $\pi(A|S,\theta)$ is parameterized by the vector $\theta$ of synaptic weights $\{w_{i,j}\}$ and biases $\{b_j\}$ for all the output neurons as defined in (\ref{eq:pot}). We now describe encoding, decoding, and learning rule.

\textbf{Encoding.} A position $S_t$ is encoded into $N_x$ spike trains, i.e., binary sequences, with duration $T$ samples, each of which is assigned to one of the neurons in the input layer of the SNN. We emphasize that the time duration $T$ is internal to the operation of the SNN, and the agent remains at time-step $t$ while waiting for the outcome of the SNN. In order to explore the trade-off between encoding complexity and accuracy, we partition the grid into $N_x$ sections, or windows, each of size $W\times W$. Each section is encoded by one input neuron, so that increasing $W$ yields a smaller SNN at the cost of a reduced resolution of state encoding. Each position $S_t$ on the grid can be described in terms of the index $s(S_t)\in\lbrace 1,...,N_x \rbrace$ of the section it belongs to, and the index $w(S_t)\in \lbrace 1,...,W^2 \rbrace$ indicating the location within the section using left-to-right and top-to-bottom ordering. Accordingly, using rate encoding, the input to the $i$th neuron is an i.i.d. spike train with probability of a spike given by
\begin{align}
\label{eq:rateencode}
     p_i= 
\begin{cases}
    p_{\text{min}}+\left(\frac{p_{\text{max}}-p_{\text{min}}}{W^2-1}\right)\left(w(S_t)-1\right),& \text{if } i=s(S_t)\\
    0,              & \text{otherwise}
\end{cases}   
\end{align}for given parameters $p_{\text{min}},\ p_{\text{max}}\in[0,1]$ and $p_{\text{max}}\geq p_{\text{min}}$.

\textbf{Decoding.} We adopt a first-to-spike decoding protocol, so that the output of the SNN directly acts as a stochastic policy, inherently sampling from the distribution $\pi(A|S,\theta)$ induced by the first-to-spike rule. If no output neuron spikes during the input presentation time $T$, no action is taken, while if multiple output neurons spike concurrently, an action is chosen from among them at random.

Given the synaptic weights and biases in vector $\theta$, the probability that the $j$th output neuron spikes first, and thus the probability that the network chooses action $A=j$, is given as $\text{Pr}(A=j)=\sum_{\tau=1}^{T} p_{\tau}(j)$, where \begin{align}
p_{\tau}(j)=\prod_{k\neq j}\prod_{\tau'=1}^{\tau}(1-\sigma(u_{k,\tau'}))\sigma(u_{j,\tau})\prod_{\tau'=1}^{\tau-1}(1-\sigma(u_{j,\tau'}))
    \label{eq:p_t}
\end{align} is the probability that the $j$th output neuron spikes first at time $\tau$, while the other neurons do not spike until time $\tau$ included. 

\textbf{Policy-gradient learning.} After an episode is generated by following the first-to-spike policy, the parameters $\theta$ are updated using the policy gradient method \cite{sutton1998reinforcement}. The gradient of the objective function (\ref{eq:value}) equals 
\begin{align}
\label{eq:stochgrad}
    \nabla_{\theta}V_{\pi}(S_0)= \text{E}_{\pi}[ V_t\nabla_{\theta}\log\pi(A_t|S_t,\theta)],
\end{align}
where $V_t=\sum_{t'=t}^{\infty} \gamma^{t'} R_{t'}$ is the discounted return from the current time-step until the end of the episode and the expectation is taken with respect to the distribution of states and actions under policy $\pi$ (see [19, Ch. 13]). The gradient in (\ref{eq:stochgrad}) can be computed as \cite{bagheri2017training}
\begin{align}
\label{eq:wgrad}
\nabla_{w_{i,k}} \log\pi_{\theta}(A_t=j|S_t,\theta)=
     \begin{cases}
      -\sum_{\tau=1}^{T}h_{\tau}\sigma(u_{k,\tau})B^{T}x_{i,\tau-\tau_s:\tau-1} & k\neq j  \\
      -\sum_{\tau=1}^{T}(h_{\tau}\sigma(u_{j,\tau})-q_{\tau})B^{T}x_{i,\tau-\tau_s:\tau-1} & k =j,
   \end{cases}
\end{align}
and
\begin{align}
\label{eq:bgrad}
\nabla_{b_{k}}\log\pi_{\theta}(A_t=j|S_t,\theta)=
      \begin{cases} 
      -\sum_{\tau=1}^{T}h_{\tau}\sigma(u_{k,\tau}) u &  k\neq j  \\
      -\sum_{\tau=1}^{T}h_{\tau}\sigma(u_{j,\tau})-q_{\tau} & k = j 
   \end{cases}
\end{align}
where \begin{align*}
h_{\tau} = \sum_{\tau'=\tau}^{T}q_{\tau'},\; \text{and}\; q_{\tau}=\dfrac{p_{\tau}}{\sum_{\tau=1}^{T}p_{\tau}}.
\end{align*} The first-to-spike policy gradient algorithm is summarized in Algorithm \ref{alg:ftsreinforce}, where the gradient (\ref{eq:stochgrad}) is approximated using Monte-Carlo sampling in each episode [19 Ch. 5].

\section{Baseline SNN Solution}
\label{sec:baselineSNN}

As a baseline solution, we consider the standard approach of converting an offline trained ANN into a deterministic IF SNN. Conversion aims at ensuring that the output spiking rates of the neurons in the SNN are proportional to the numerical values output by the corresponding neurons in the ANN \cite{diehl2015fast}. 

\textbf{IF neuron.} The spiking behavior of an IF neuron is determined by an internal membrane potential defined as in (\ref{eq:pot}) with the key differences that: (\emph{i}) the synaptic kernels are perfect integrators, that is, they are written as $\alpha_{i,j}=w_{i,j}1$, where $w_{i,j}$ is a trainable synaptic weight and $1$ is an all-one vector of $T$ elements; and (\emph{ii}) the neuron spikes deterministically when the membrane potential is positive, so that parameter $b_j$ plays the role of negative threshold.

\textbf{Training of the ANN and Conversion into an IF SNN.} A two-layer ANN with four ReLU output units is trained by using the SARSA learning rule with $\epsilon$-greedy action selection in order to approximate the action-value function of the optimal policy \cite{sutton1998reinforcement}. The input to each neuron $i$ in the first layer of the ANN during training is given by the probability value, or spiking rate, $p_i$ defined in (\ref{eq:rateencode}), which encodes the environment state. Each output neuron of the ANN encodes the estimated value, i.e., the estimated long-term discounted average reward, of one of the four actions for the given input state and the action with the maximum value is chosen (under $\epsilon$-greedy action choices) with probability $\epsilon$. The ANN can then be directly converted into a two-layer IF SNN with the same architecture using the state-of-the-art methods proposed in \cite{diehl2015fast}, to which we refer for details. The converted SNN is used by means of rate decoding: the number of spikes output by each neuron in the second layer is used as a measure of the value of the corresponding action. We emphasize that, unlike in the proposed solution, the resulting (deterministic) IF SNN does not provide a random policy but rather a deterministic action-value function approximator.

\section{Results and Discussion}
\label{sec:results}
In this section, we provide numerical results for the grid world example described in Sec. \ref{sec:prob def} with $M=7$, $N=10$, $S_0$ and $S^{\text{G}}$ at positions (4,1) and (4,8) on the grid respectively, `wind' level per columns defined by the values $\omega_n$ indicated in Fig. \ref{fig:optimal policy}(b), and $K_s=\tau_s$ for all simulations. Throughout, we set $p_\text{min}=0.5$ and $p_\text{max}=1$ for encoding in the spike domain and a learning schedule, $\eta_{i} = (\eta_{i-1})/(1-k(i-1))$ with $\eta_0=10^{-2}$. Training is done for 25 epochs of 1000 iterations each, with 500 test episodes to evaluate the performance of the policy after each epoch. Hyper-parameters for the SARSA algorithm to be used as described in the previous section are selected as recommended in \cite{mnih2013playing,lin1993reinforcement}.

\begin{figure}[!th]
    \centering
    \includegraphics[scale = 0.5,trim={0.4in 0.15in 0.95in 7.1in},clip]{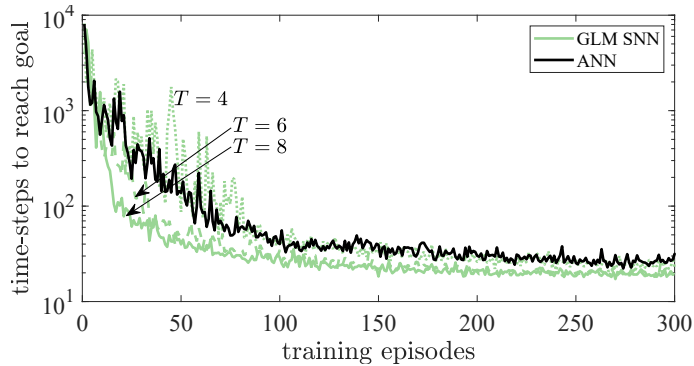}
    \caption{Number of time-steps needed to reach the goal state as a function of the training episodes for the proposed GLM SNN and the reference ANN strategies.}
\label{fig:res}
\end{figure}

Apart from the IF SNN solution described in the previous section, we also use as reference, the performance of an ANN trained using the same policy gradient approach as in Algorithm 1 and having the same two-layer architecture as the proposed SNN. In particular, the input to each input neuron $i$ of the ANN is again given by the probability $p_i$ defined in (\ref{eq:rateencode}), while the output is given by a softmax non-linearity. The output hence encodes the probability of selecting each action. It is noted that, despite having the same architecture, the ANN has fewer parameters than the proposed first-to-spike SNN: while the SNN has $K_s$  parameters for each synapse given its capability to carry out temporal processing, the ANN has conventionally a single synaptic weight per synapse. This reference method is labeled as ``ANN" in the figures.

We start by considering the convergence of the learning process along the training episodes in terms of number of time-steps to reach the goal state. To this end, in Fig. \ref{fig:res}, we plot the performance, averaged over the 25 training epochs, of the first-to-spike SNN policy with different values of input presentation duration $T$ and GLM parameters $K_s=\tau_s=4$, as well as that of the reference ANN introduced above, both using encoding window size $W=1$, and hence $N_x=70$ input neurons. We do not show the performance of the IF SNN since this solution carries out offline learning (see Sec. IV). The probabilistic SNN policy is seen to learn more quickly how to reach the goal point in fewer time-steps as $T$ is increased. This improvement stems from the proportional increase in the number of input spikes that can be processed by the SNN, enhancing the accuracy of input encoding. It is also interesting to observe that the ANN strategy is outperformed by the first-to-spike SNN policy. As discussed, this is due to the capability of the SNN to learn synaptic kernels via its additional  weights. 
\begin{figure}[!th]
     \centering
    \includegraphics[scale=0.5,trim={0.6in 0.15in 0.92in 6.8in},clip]{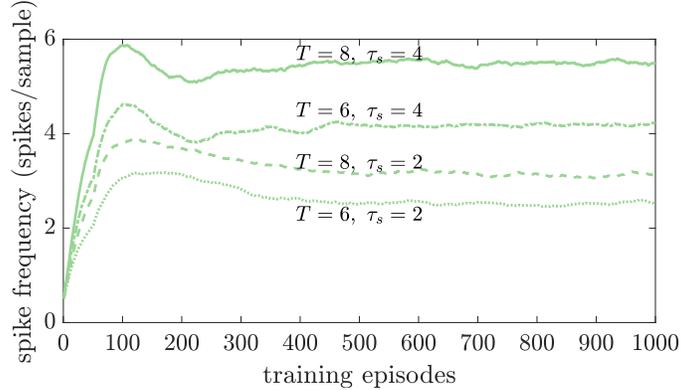}
    \caption{Average spike frequency over the training episodes for the GLM SNN policy.}
    \label{fig:spikesperchoice}
\end{figure}

We further investigate the behavior of the first-to-spike SNN during training in Fig. \ref{fig:spikesperchoice}, which plots the spike frequency as a function of the training episodes. The initially very low spike frequency can be interpreted as an exploration phase, where the network makes mostly random action choices by largely neglecting the input spikes. The spike frequency then increases as the SNN learns while exploring effective actions dictated by the first-to-spike rule. Finally, after the first one hundred episodes, the SNN learns to exploit optimal actions, hence reducing the number of observed spikes necessary to fire the neuron corresponding to the optimized action.

\begin{figure}[!ht]
    \centering
    \includegraphics[scale=0.5,trim={0.5in 0.3in 1in 3.75in},clip]{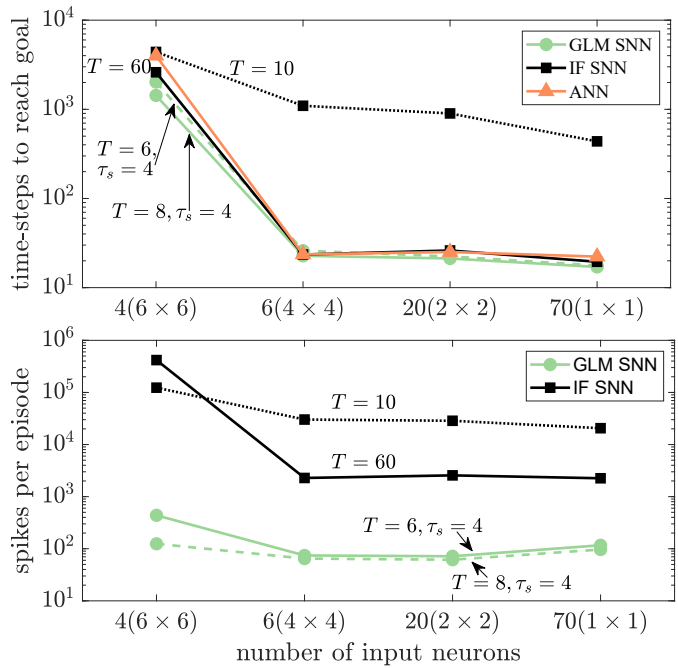}
    \caption{Average number of time-steps to reach goal (top) and average number of total spikes per sample (bottom) in the test episodes as a function of the number of input neurons $N_x$ (also indicated is the size of the $W\times W$ encoding window) for GLM SNN and IF SNN.}
    \label{fig:perfVspikef}
\end{figure}
We now turn to the performance evaluated after training. Here we consider also the performance of the conventional IF SNN trained offline as described in Sec. IV. We first analyze the impact of using coarser state encodings as defined by the encoding window size $W$. Considering only test episodes, Fig. \ref{fig:perfVspikef} plots the number of time-steps to reach the goal (top) and the total number of spikes per episode across the network (bottom), as a function of the number of input neurons, or equivalently of $W$. For all schemes, it is seen that, as long as the window size is no larger than $W=4$ and $T$ is large enough for the SNN-based strategies, no significant increase of time-steps to reach the goal is incurred. Importantly, the IF SNN is observed to require $10\times$ the presentation time and more than $10\times$ the number of spikes per episode of the first-to-spike SNN to achieve the same performance.

\begin{figure}[!th]
    \centering
   \includegraphics[scale=0.5, trim = {0.4in 0.15in 0.98in 7.05in}, clip]{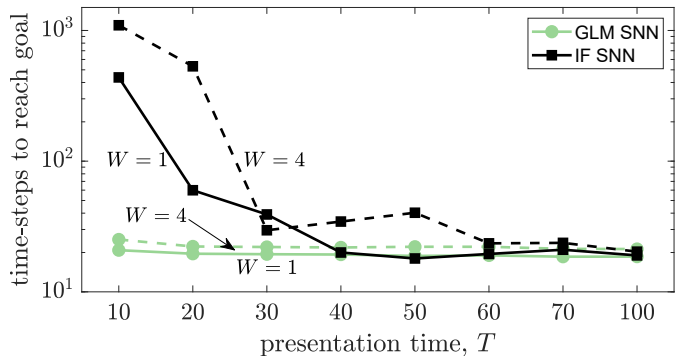}
    \caption{Average number of time-steps to reach goal versus the presentation time $T$ for GLM SNN and IF SNN ($\tau_s=6, K_s=1$).}
    \label{fig:perfVT}
\end{figure}

The test performance comparison between first-to-spike SNN and IF SNN is further studied in Fig. \ref{fig:perfVT}, which varies the presentation time $T$. In order to discount the advantages of the first-to-spike SNN due to its larger number of synaptic parameters, we set here $K_s=1$, thus reducing the number of synaptic parameters to 1 as for the IF SNN. Fig. \ref{fig:perfVT} shows that the gains of the proposed policy are to be largely ascribed to its decision rule learned based on first-to-spike decoding. In contrast, the IF SNN uses conventional rate decoding, which requires a larger value of $T$ in order to obtain a sufficiently good estimate of the value of each state via the spiking rates of the corresponding output neurons.
\section{Conclusions}
This paper has proposed a policy gradient-based online learning strategy for a first-to-spike spiking neural network (SNN). As compared to a conventional approach based on offline learning and conversion of a second generation  artificial neural network  (ANN) to an integrate-and-fire (IF) SNN with rate decoding, the proposed approach was seen to yield a reduction in presentation time and number of spikes by more than $10\times$ in a standard windy grid world example. Thanks to the larger number of trainable parameters associated with each synapse, which enables optimization of the synaptic kernels, performance gains were also observed with respect to a conventional ANN with the same architecture that was trained online using policy gradient. 

\bibliographystyle{IEEEbib}
\bibliography{references}

\end{document}